\documentclass[nonacm,screen]{acmart}
\usepackage[british]{babel}

\usepackage{array} %
\usepackage{multirow}

\begin{document}

\title{Reproducibility in Evolutionary Computation}

\author{Manuel L{\'o}pez-Ib{\'a}{\~n}ez}
\email{manuel.lopez-ibanez@uma.es}
\orcid{0000-0001-9974-1295}
\affiliation{%
  \institution{University of M{\'a}laga}
  \streetaddress{Bulevar Louis Pasteur, 35}
  \city{M{\'a}laga}
  \country{Spain}
  \postcode{29071}
}

\author{Juergen Branke}
\email{juergen.branke@wbs.ac.uk}
\orcid{0000-0002-4343-5878}
\affiliation{%
  \institution{University of Warwick}
  \streetaddress{Gibbet Hill Road}
  \city{Coventry}
  \country{UK}
  \postcode{CV4 7AL}
}

\author{Lu\'is Paquete}
\email{paquete@uc.pt}
\affiliation{%
  \institution{University of Coimbra, CISUC, Department of Informatics Engineering}
  \streetaddress{Polo II, Pinhal de Marrocos}
  \city{Coimbra}
  \country{Portugal}
  \postcode{3020-290}
}

\begin{abstract}
  Experimental studies are prevalent in Evolutionary Computation (EC), and
  concerns about the reproducibility and replicability of such studies have
  increased in recent times, reflecting similar concerns in other scientific
  fields. In this article, we discuss, within the context of EC, the different types of reproducibility and suggest a classification that refines the badge system of the Association of Computing Machinery (ACM) adopted by ACM Transactions on Evolutionary Learning and Optimization (\url{https://dlnext.acm.org/journal/telo}).
We identify
  cultural and technical obstacles to reproducibility in the EC
  field. Finally, we provide guidelines and suggest tools that may
  help to overcome some of these reproducibility obstacles.
\end{abstract}

\keywords{Evolutionary Computation, Reproducibility, Empirical study, Benchmarking}

\maketitle

\section{Introduction}
\label{sec:intro}

As in many other fields of Computer Science,
most of the published research in Evolutionary Computation (EC) relies on experiments to justify their
conclusions.
The ability of reaching similar conclusions by repeating an experiment
performed by other researchers is the only way a research community can reach a
consensus
on   an empirical claim until a mathematical proof is discovered.
From an engineering perspective, the assumption that experimental findings hold
under similar conditions is essential for making sound decisions and predicting
their outcomes when tackling a real-world problem. 

The ``reproducibility crisis'' refers to the realisation that
many experimental findings described in peer-reviewed scientific publications
cannot be reproduced, because e.g. %
 they lack the necessary data, they lack enough details to repeat the
experiment or repeating the experiment leads to different
conclusions. Despite its strong mathematical basis, Computer Science (CS) also
shows signs of suffering such a crisis~\citep{CocDraBes2020threats,FonTag2020repro,GunGilAha2018repro}.
 EC is by no
means an exception. In fact, as we will discuss later, particular challenges of reproducibility in EC arise from the stochastic nature of the algorithms.

Although concerns about reproducibility in randomised search heuristics have existed for a long time, see, for example, \citet{GenGraMac1997hownotto}, \citet{Johnson2002}, and \citet{EibJel2002critical}, only recently we have reached a critical point that is leading to changes in journal policies and research practices.
  The goal of this paper is to discuss reproducibility in the context
  of EC (and randomised search heuristics in general). We review the abundant
  research on reproducibility from other fields and adapt it, when pertinent, to
  the EC context.
In Section~\ref{sec:why}, we explain what reproducibility means in the context
of EC and argue that reproducibility is as relevant in EC as in any other
sub-field of CS, both from a scientific and from
 an engineering perspective.
In Section~\ref{sec:terminology}, we discuss, in
the context of EC, two key concepts that arise when discussing reproducibility,
the notions of \emph{artifact} and \emph{measurement}. We also  review the terminology adopted by ACM \citep{ACM2020badging_v1_1} and
others to formally distinguish between different levels of reproducibility, and
propose a refinement  that classifies 
  reproducibility studies in EC according to the factors that
  are varied in the study with respect to the original work.
  We discuss in
Section~\ref{sec:obstacles} some of the cultural and technical obstacles to
ensuring reproducibility in EC. In Section~\ref{sec:guidelines}, we suggest
guidelines and tools that may help overcome some of those obstacles. Finally,
we conclude in Section~\ref{sec:conclusions} with an overall discussion of the
state of reproducibility in EC and point out future directions to further
understand and improve reproducibility.

\section{Why is reproducibility in EC important ?}
\label{sec:why}

\subsection{Falsifiability and community consensus}
Evolutionary Computation (EC), in much the same way as Computer
Science~\citep{Wegner76research}, can be seen as a three-fold discipline: it is
\emph{mathematical} since it is concerned with the formal properties of
abstract structures; it is also \emph{scientific} since it is concerned with
the empirical study of a particular class of phenomena; and it is
\emph{engineering} since it is concerned with the effective design of tools
that have social and commercial impact in the real world. 
Despite major advances in the Theory of EC,  the dynamics of practical EC algorithms applied to non-trivial problems  are still too 
complex to be analysed using only mathematical arguments. As a result, a majority of EC research relies on empirical studies. Thus, in the following, we focus on the scientific (empirical) and engineering perspectives.

In empirical sciences, the body of knowledge is built by following the principles of the scientific method:

\begin{enumerate}
\item Observe a phenomenon, e.g. EAX crossover appears to have the capacity for local optimisation in the travelling salesman problem (TSP)~\citep{NagKob1997}.
\item Construct a hypothesis, e.g. a specific evolutionary algorithm (EA) converges faster to the optimal solution for a particular class of TSP instances when using EAX  than when using the other known crossovers for the TSP.
\item Conduct an experiment, e.g. measuring the performance of the EA using EAX and alternative crossovers on a number of TSP instances.
\item Analyse and draw a conclusion on whether the experiment supports the hypothesis and, hence, it is provisionally accepted, or not, hence, it is \emph{falsified}.
  
\end{enumerate}

A cornerstone of the scientific method is the notion of
\emph{falsifiability}, i.e. a scientific hypothesis must be testable
empirically and, possibly, falsifiable.  For example, the
statement ``\emph{There are problems for which Evolutionary Algorithms are the best optimisation
  methods possible}'' is not falsifiable (by evidence), not only because of the
vagueness of terms such as ``best'' and ``Evolutionary Algorithm'', but also because we may never
know all possible optimisation methods nor all possible problems.
However, the statement ``\emph{in crossover operators for the traveling salesman
problem, the trade-off between the ratio of edges inherited by offsprings from
parents and the variety of offsprings is important for generating large number
of improved offsprings}''~\citep{NagKob1999analysis} is falsifiable by evidence.

On the other hand, research in EC very often takes an engineering perspective,
e.g. when  
comparing different EC algorithms to solve a particular problem. 
As in other engineering disciplines, a researcher has to go through the
following steps:
\begin{enumerate}
\item Specify requirements, e.g. an EA that outperforms the LKH algorithm in
  finding solutions less than $1\%$ from the optimum on instances of the
  TSP with up to 200,000 cities~\citep{NagKob2013}.
\item Design a solution, e.g. a particular type of EA that uses a novel
  crossover.
\item Conduct an experiment, which will often involve implementing a reasonably efficient prototype, careful parameter tuning and benchmarking the prototype against a competitor.
  \item Analyse and draw a conclusion on whether the benchmarking results provide evidence that the solution meets the requirements.
  
\end{enumerate}

There are clear parallels between the scientific and engineering
perspectives. Moreover, engineering requirements can also be recast as
scientific hypotheses. A major difference between the scientific and engineering
perspectives is that the latter is mostly concerned with demonstrating
performance differences between realistic algorithmic implementations on
practical problems under the requirements specification, whereas the former
is concerned with confirming hypotheses on abstract models of the real-world
that may lead to general principles.

From both scientific and engineering perspectives, experiments that are reproducible and falsifiable by others are a prerequisite to reach a consensus in the research community and building a body of knowledge about working principles of EC. Such  ``laws of
qualitative structure''~\citep{NewSim1976cacm} are qualitative hypotheses
that are accepted by the research community until sufficient empirical evidence
arises to falsify them, e.g.  the generally accepted hypothesis that the search
space of the travelling salesman problem (TSP) has a ``big valley''
structure~\citep{BoeKahMud1994}.

\subsection{Building on the work of others}

Scientific progress is a collaborative effort. Most new research results build on previous research results. The first step to improving an algorithm is to reproduce the previous results. In this sense, reproducibility 
facilitates (or even is a prerequisite of) scientific progress. If reproducing previous results is easy because the research has been published in a reproducible format, it saves researchers a lot of time, allowing the community to quickly absorb new results, speeding up scientific progress as well as the transfer of new ideas into practical applications.

\subsection{Quality control and error correction}

In a provocative paper entitled "Why most published research findings are false", \citet{Ioa2005why} suggests that much of published research results cannot be trusted. A recent survey in Nature \cite{Baker2016reprod} revealed that more than 70\% of researchers have previously failed in an attempt to reproduce another researcher's experiments, and over 50\% have even failed to reproduce one of their own previous results.
This is generally not due to researchers deliberately falsifying their results, but more often a result of the publication culture, researcher ignorance, or confirmation bias. 

A well-documented bias against publishing negative results \cite{Fanelli2012} together with a culture that rewards scientists chiefly on quantity of publications incentivises non-reproducible 
research~\citep{GriBauIoa2018modeling}.

Researchers often lack sufficient expertise in statistics and unknowingly use improper statistical tests, insufficient sample sizes~\citep{CamWan2020sample}, or manipulations of the
experimental conditions that alter (unintentionally or deliberately) the
statistical significance of results, e.g. p-hacking~\citep{SimNelSim2011phack,CocDraBes2020threats} and hypothesising after results are known  (HARKing)~\citep{Kerr1998harking}.

In particular since research in EC is often framed as competitive
testing~\citep{Hoo1996joh}, there is also a bias in the effort spent by
researchers in verifying their experimental setup and code. If a researcher hypothesises  that
a new mutation operator should work well on a particular problem, and
experiments show poor performance, they are likely to carefully check their
code and experimental setup to make sure that this unexpectedly poor
performance is not due to an error. On the other hand, if results are very
positive, they are less likely to suspect a problem and thus spend much less
time verifying their code and experimental setup. As a result, errors that lead
to poor performance are usually corrected, whereas errors that lead to apparent
but false good performance are often not detected and are published. A similar
bias, but in the opposite direction, may be true for competing
algorithms. There, an error that leads to poor performance of an existing
algorithm relative to the author's newly proposed algorithm risks not being
detected because it supports the author's presumption that their new algorithm
is better. Even if the code is available but the bug only shows up on new
problem instances or experimental conditions, there is little incentive to
investigate the reasons behind the poor performance of a benchmark algorithm.
\citet{Bro2015emo} reports the illustrative case of a bug in one implementation
of an algorithm affecting published results and how the bug has propagated to
many other software packages due to the lack of independent implementations,
potentially affecting the results of hundreds of published papers.

However, even though there is evidence that a significant proportion of published research results are wrong, and many researchers probably have experienced challenges in reproducing published results~\citep{SorArnPal2017}, the number of published corrections is negligible. A search on Scopus (November 2020)
reveals that out of 2484 papers published in the journals \emph{IEEE Transactions on Evolutionary Computation}, \emph{Evolutionary Computation}, and \emph{Swarm Intelligence and Evolutionary Computation}, only 8 were Errata.
As a consequence, a lot of effort is potentially 
wasted by many research groups who independently attempt to reproduce results and fail, before the rumor somehow spreads and people accept that certain results are not reproducible.
A proper research culture  where reproducibility is regularly attempted and also negative results are published could significantly speed up scientific progress.

\section{Terminology}
\label{sec:terminology}

Informally, the terms reproducibility and replicability are often used to
describe various concepts related to being able to confirm or falsify a
hypothesis by repeating an experiment. More formally, those terms often denote
various degrees of reproducibility and, unfortunately, not always consistently, since different communities use different terminologies. For a historical
perspective on terminology see \citet{Ples2018repro}. In our paper, we use the
term ``reproducibility'' when addressing the general topic. When discussing
specific degrees of reproducibility, we mostly follow the terminology used by
ACM~\citep{ACM2020badging_v1_1} with a further refinement presented later in this Section.\footnote{On August 24th 2020, ACM swapped the
  definitions of ``Reproducibility'' and ``Replicability'' to match the
  terminology proposed by \citet{ClaKar1992electronic}.}
The ACM terminology relies on the concepts of ``artifact'' and ``measurement'':

\paragraph*{Artifact} ``A digital object that was either created by the authors
to be used as part of the study or generated by the experiment
itself''~\citep{ACM2020badging_v1_1}. Examples of artifacts in the context of
EC would be complete implementations of algorithms, either in source code or
executable form; data or code required to fully specify problem instances or
benchmark functions, e.g. files containing distance matrices for the traveling
salesman problem, a software library of continuous benchmark functions or a
simulation software needed for evaluating solutions; raw data measured during
the experiment and used for validating the hypothesis, e.g. measurements of
solution quality, counts of objective function evaluations, iterations, steps,
or computation times; and any scripts required to process the raw measurements
and calculate the statistics or visualisations that justify the conclusions of the
experiment.

\paragraph*{Measurement} The term ``measurement'' is used in analogy to physical experiments. For computer science, a measurement is the raw data (objective function values, runtimes, etc.) that results from an experiment. In EC, instead of the actual measurements, it is common to report summary statistics such as means and standard errors.
As discussed
by \citet{McG2012guide}, the measurements
taken should be appropriate to the level of abstraction being
studied. For example, computational effort may be measured as cycles, seconds,
function evaluations or iteration counters.

\medskip{}
Based on the above concepts of artifact and measurement, the ACM defines the following
terms~\citep{ACM2020badging_v1_1}:
\paragraph*{Repeatability}
``The measurement can be obtained with stated precision by the same team using the same measurement procedure, the same measuring system, under the same operating conditions, in the same location on multiple trials.'' \emph{(Same team, same experimental setup)}.

\paragraph*{Reproducibility} %
``The measurement can be obtained with stated precision by a different team using the same measurement procedure, the same measuring system, under the same operating conditions, in the same or a different location on multiple trials.'' \emph{(Different team, same experimental setup)}.

\paragraph*{Replicability} %
``The measurement can be obtained with stated precision by a different team, a different measuring system, in a different location on multiple trials.'' \emph{(Different team, different experimental setup)}.

\medskip{}
In the context of EC, repeatability means that the authors of a publication can
reliably perform multiple times their own experiments and get the same result
up to their own stated precision. Reproducibility means that independent
researchers can reliably perform multiple times the experiments described by
the publication using the artifacts provided by the original authors and the
same computational environment or a similar one, and get the same result up to
the stated precision. Finally, replicability means that independent researchers can
reliably perform multiple times the experiments using independently developed
artifacts on a different computational environment and get the same result up
to the stated precision.

According to the ACM classification, the main distinction between
reproducibility and replicability is that, in case of reproducibility, the
original artifacts are re-used, while for replicability another group has to
independently generate the necessary artifacts.  However, we believe that there
are more dimensions to reproducibility, especially in evolutionary computation,
where algorithms are randomised, parameterised, and results based on benchmark
problems. What should be kept fixed and what should change to assess either
reproducibility or replicability?

Following classical statistical terminology~\citep{ChiGoe2010}, we make a distinction between two
types of experimental factors: \emph{random effect factors} and \emph{fixed
  effect factors}. A random factor has many possible values and the experimental conclusion of a paper applies to a certain range or distribution, but the
experiment only evaluates a random sample of values. A fixed factor may
also have many possible values, but the experiment only evaluates specific
values chosen by the experimenter and the claim in the paper is only supported
for those specific values. A typical random factor in EC is the random seed of
a stochastic algorithm, even though most computer experiments are not truly random.
A typical fixed factor would be an algorithmic parameter. Whether a factor is
treated as random or as fixed is typically decided by the experimenter,
depending on the claim that the experiment will aim to support. In some cases,
a factor must be fixed because there is no known unbiased way to sample its
values.  This is often the case for benchmark problem instances---e.g. it is
not clear how to sample from the space of all ``interesting'' real-valued
functions---or only few real-world instances are available for a particular
application. If they are selected by the experimenter, then they are treated as
a fixed factor and the experiment only directly supports claims regarding those
specific instances, although the author may hypothesise about a wider
applicability. If the problem instances are randomly generated or selected from
a larger class of instances, then they are treated as a random factor, and the
paper can make statistical inferences about the larger class.

We suggest to consider the following three dimensions of reproducibility:
\begin{enumerate}
\item \emph{Artifacts:} Re-use of the original artifacts should allow to repeat the exact same experiments as described in the original publication. However, it bears the risk of also repeating the exact same mistakes in case the original code or data contained errors. Having the artifacts re-created by another group reduces the risk of  errors being repeated, and also confirms that all information required to re-create the artifacts is contained in the manuscript. We extend ACM's definition of artifact beyond pure digital objects and suggest that, in some cases, the entire computational environment, and even the hardware, used in the original experiment may be provided as artifacts in the form of virtual machines, ``containers'' or access to cloud platforms (see Section~\ref{sec:guide-ensuring}).
\item \emph{Random factors:} In the presence of random factors in an experiment,
  repeating exactly the same computation would require using exactly the same
  values of the random factors, e.g. same random seeds. However, one would
  expect that the claims of the paper hold after resampling the values of the
  random factors. Of course, such claims would need to be expressed in
  statistical terms to determine whether the results are equivalent.

\item \emph{Fixed factors:} Unless somehow randomised, fixed factors in EC typically
  include test problems, parameter settings, computational budget,
  etc. Strictly speaking, the hypothesis supported by the experiment will only apply to
  the specific values tested.  Changing these values (or converting them to a
  random factor) will test whether the claims of the paper generalise also to
  other values and would go beyond just replication of the experiments in the
  paper. In some cases, the experiment specifies a reproducible procedure to
  randomise or unambiguously determine the values of a factor, for example, for
  deciding parameter values. In those cases, the procedure itself becomes the
  experimental factor, either random or fixed.
\end{enumerate}

The typical combinations of these dimensions in a reproducibility study,
together with a suggested label, are summarised in Table~\ref{tab:terms}.
In a \emph{repeatability} study, every dimension is exactly as in the original
experiment. This could be useful to assess that the original results are indeed
obtainable, but may be only feasible for the original authors or require
access to the original computational environment. A \emph{reproducibility}
study (in the narrow sense) would vary the stochastic aspects of the
experiment, i.e. the random factors, but re-use as much as possible the
original artifacts, possibly including the computational environment if
provided as an artifact, and values of fixed factors. At this level, we cannot
expect to obtain exactly the same results as the original experiment.  What is
being evaluated is the statistical robustness of the conclusions reached.  At a
third level we find \emph{replicability} studies, where the goal is to reach
the same conclusion as the original experiment but with independently developed
artifacts. Such a study would evaluate how much the conclusions depend on the
particular artifacts and/or computational environment. As in the previous
level, random factors must be varied to properly evaluate the statistical
robustness of the claims, thus there is no point in re-using the original
random seeds at this level.  A further level concerns the
\emph{generalisability} of claims of the paper to other values of the fixed
factors. Generalisability goes beyond the claims supported by the
experiment. For example, although the conclusions of the paper may be true for
the problem instances (or instance generator) evaluated, they would be more
interesting if they extend to other problem instances. The sensitivity of the
algorithm's performance to particular parameter settings would also be an
example of generalisability. In generalisability studies that use independently
developed artifacts, it is a good idea to conduct first a replicability study
so that, if conclusions are different from the ones in the original experiment,
this discrepancy can be properly attributed to the changes in fixed factors or
in artifacts.
 
Studies that fall between levels are possible. For example, a study that
re-uses some of the original artifacts, such as the implementation of the
algorithms, while evaluating the results in a new computational environment,
would fall closer to reproducibility than replicability. Similarly, if claims
of the paper rely on specific aspects (implementation language, hardware
and third-party software capabilities, etc), these become fixed factors rather than
artifacts and, thus, varying them would be closer to a generalisability study
than a replicability one.

\begin{table}
  \centering
  \caption{Proposed classification of reproducibility studies.}
  \label{tab:terms}
  \begin{tabular*}{\linewidth}{llll>{\raggedright\arraybackslash}p{25em}}
    \toprule
\bf Label        &\bf\shortstack{Artifacts} & \bf\shortstack[l]{Random\\ factors} & \bf\shortstack[l]{Fixed\\ factors} &  \bf Purpose of the study                                                                                                                 \\\midrule
Repeatability    &Original                  & Original                          & Original                         &  Exactly repeat the original experiment, generating precisely the same results.                                        \\\\
Reproducibility  &Original                  & New                               & Original                         &  Test whether the original results were dependent on specific values of random factors and, hence, only a statistical anomaly. \\\\
Replicability    &New                       & New                               & Original                         &  Test whether it is possible to independently reach the same conclusion without relying on original artifacts.         \\\\
Generalisability & \multirow{2}{*}{\shortstack[l]{Original\\ or New}}                       & New                               & New                              &  Test whether the conclusion extends beyond the experimental setup of the original paper.         When new artifacts are used, generalisability should come after a replicability study.                             \\\bottomrule
  \end{tabular*}
\end{table}

Ideally, all published experiments should be replicable, and some argue that a pure repeatability study does not generate additional evidence for a paper's claims and therefore may not be worthwhile \cite{Dru2009replicability}.
However, evaluating the repeatability and
reproducibility of an experimental study is typically less demanding and may be
taken as a precondition before attempting replication.
In any case, we consider it important that reproducibility studies are specific about what level of reproducibility is attempted. We also suggest that if an attempt to reproduce results fails to generalise to other values of fixed factors (e.g. test problems or parameter settings), it should be attempted with the same test problems and parameter settings, and if this fails too, it should be attempted with the original artifacts and, if possible,  same random seeds. This would allow tracing back the cause of  a reproducibility problem. Hence, while repeatability and reproducibility  are not the end goal, they are still important to learn from and facilitate replicability and generalisability studies.

For completeness, we would like to mention two more terminologies that classify levels of reproducibility. First, the Turing Way project~\citep{TuringWay2019} funded by The Alan Turing Institute in the UK distinguishes reproducibility (same analysis performed on the same dataset consistently produces the same answer), replicability (same analysis performed on different datasets produces qualitatively similar answers), robustness (different analysis applied to the same dataset produces a qualitatively similar answer to the same question) and generalisability (the combination of replicability and robustness). This classification has been adopted by a segment of the Machine Learning community~\citep{PinVinSin2020improving}.
Second, \citet{Sto2014edge} makes a distinction between \emph{empirical
  reproducibility}, which is the concept of reproducibility that arises in
natural sciences, i.e. being able to repeat an experiment following the
details published and obtain a similar conclusion by using different artifacts,
since artifacts cannot be copied in the natural sciences; \emph{computational
  reproducibility}, which relates to the availability of the code, data and all
details of the implementation and experimental setup that allow obtaining the
published results; and, lastly, \emph{statistical reproducibility}, which is
concerned with validating the results of repeated experiments by means of
statistical assessment. 

Nevertheless, the above definitions do not fully specify what details define
the experimental setup (or operating conditions). Completely replicating the
exact conditions of the original experiment may be impossible even by the
original authors, e.g. the original hardware may not be available anymore, the
load of the computational system may have influenced the measurements, the
precise version of some software libraries may be unknown, some sources of
randomness may not be repeatable, etc. In that case, the experimental setup may
refer only to the details that the original authors consider relevant for their
experiment. Alternatively, one may give up on repeatability and reproducibility
as long as replicability is achievable, which does not mean that the latter is
easier to achieve than the former. Indeed, replicability requires high-level
descriptions of artifacts with enough detail to enable their independent
development and a careful choice of measurements, their stated precision and
confidence levels that allow other researchers to unequivocally conclude
whether a replication attempt falsifies the original experiment.

\section{Obstacles to reproducibility}\label{sec:obstacles}

Despite the obvious benefits or reproducibility studies, very few such studies are published in EC.
In this section, we try to explain the low number of reproducibility studies by discussing cultural and technical obstacles.

\subsection{Cultural obstacles}

A key reason for the low number of reproducibility studies is simply that the
current ``publish or perish'' culture does not encourage it.
Neither has the author of a scientific paper enough incentives to facilitate reproducibility
studies, nor have other scientists enough incentives to conduct reproducibility studies.

\paragraph{Disincentives to artifact publication} Reproducibility studies would be
greatly facilitated if authors would make their artifacts
available and accessible for others to be easily re-used, which means that authors would have to learn about and apply
standard principles of software engineering such as proper documentation,
modularity, version control, testing and maintenance. With reputation and
career prospects closely linked to the number of publications, this additional
effort is not obviously beneficial to the individual,\footnote{%
  Although a recent study found a small positive correlation between linking to artifacts in a paper and its scientific impact in terms of citations \cite{HeuNieKru2020publish}.} %
and thus researchers
rather invest their time in publishing more papers than in making artifacts
available and accessible.  Besides the additional effort required, the
publication of artifacts increases the chances of error detection, and thus may
increase chances of the paper being rejected (if the artifacts are checked
before publication), having to publish errata, or even to retract a paper.

In principle, the lack of intrinsic incentives could be counter-balanced by top
journals requiring the publication of artifacts \emph{prior} to publication of
the paper. This would not only lead to a larger number of artifacts being
available, but also raise the quality of the artifacts provided, as reviewers
would be less inclined to review and accept poor quality artifacts. There is
some evidence that journal policies are frequently ignored and marginally
effective if they merely ``encourage'' authors to make their artifacts
available or only ``require'' them post-publication~\citep{StoSeiZha2018pnas}.
Nevertheless, we are not aware of any journal or conference in the EC field
that requires artifact publication. And even conferences with an
  explicit reproducibility checklist do not absolutely request artifacts upon
  submission~\citep{AAAI2021checklist,IJCAI2021checklist}.  In any case, code
review places a significant additional workload onto reviewers and a peer
review system that is already stretched to the limit. Moreover, the time constraints for conference publications would not allow
for additional code review.
It can therefore be expected that only few top publications would be able to
provide such a service. As a result, very few published papers, even in major
journals, provide a complete set of artifacts.

\paragraph{Difficulty of publishing a reproducibility study} Conducting a reproducibility study is also not incentivised, as publishing the results may be challenging \citep{SorArnPal2017}. If the experiments confirm the results from the original paper, the knowledge gained may be considered marginal. On the other hand, if the experiments fail to validate previous work,
the results of the original publication stand against the results of the reproducibility study, and the question arises whether there is a problem with the original paper, a problem with the reproducibility study, or whether the difference is simply due to statistical uncertainty.
It is difficult to convince reviewers that the new study is more reliable than the old one. An independent third party, or a collaboration between the authors of the original paper and the team that tried to reproduce the results, may be required to explain the observed difference. Hence, rather than spending the effort on reproducibility studies that are difficult to publish, scientists are incentivised to develop new algorithms and publish new results.

\paragraph{Insufficient description} The reluctance of authors to publish and
properly document their artifacts further compounds the disincentive for
reproducibility studies. Without artifacts, direct reproducibility is
impossible.  
Often, the description given in a paper is unintentionally ambiguous or insufficient to re-implement the precise algorithm. Indeed, the page limit imposed by some journals often necessitates omitting some details. This stresses the importance of making the original source code available. Even ``obsolete'' code, which  can no longer be run because the compiler or hardware needed are no longer available, can help to resolve ambiguities and fill in details missing in the paper.

But even if the artifacts are available when the paper is published,  they may not match the required standards for
reproducibility: the steps to reproduce the results are not fully documented,
the artifacts require precise versions of additional software not provided nor
documented, the download link to the artifacts has become unavailable since the paper was published, etc.
It  also happens  that the artifacts provided do not match the
description in the paper, i.e. either the algorithm described in the paper is
not the one actually used in the experiments or the results shown in the paper
correspond to a different version of the artifacts provided.

\paragraph{Mistakes perceived negatively} Even though everyone occasionally makes mistakes and discussing them openly would be beneficial to the community, errors are culturally disdained. The author of a study may feel uneasy having to admit a mistake in a published paper, and the scientist who conducted a reproducibility study may feel uneasy about challenging the authors of the original study. As a result, even if someone has attempted to reproduce a scientific study, the results are rarely published.

\subsection{Technical obstacles}

\paragraph{Intellectual property}
Concerns about licensing, privacy and commercially sensitive information may be
legitimate obstacles for making artifacts (source code or data) publicly
available~\citep{FonTag2020repro}. Although it may be tempting to make
artifacts available only to reviewers under some type of nondisclosure
agreement~\citep{StoMcNutBai2016enhancing,Her2015toms}, such an arrangement
does not actually improve reproducibility. 

\paragraph{Binary-only artifacts}
Similarly, publishing artifacts in executable form instead of source code does
not increase reproducibility. One might think that being able to reproduce the
results by having the algorithm in executable form is better than not being
able to reproduce the results.  However, such argument misunderstands the
ultimate purpose of reproducibility, which is to be able to understand in
detail how the published results were produced and whether they match the
description in the paper. Therefore, if we have to choose, even obsolete source code, in the sense discussed above, is better than ``working'' black-box object code.

\paragraph{Unreproducible Computational environment}
Although lab conditions in computer science are very controlled, conclusions may depend on details of the computational environment such as the compiler, the hardware, or specific libraries~\citep{BocFawVal2018performance}.

\paragraph{Computational resources}
A more challenging obstacle arises when the time or computational resources
required to reproduce an experiment are prohibitively large. It is not unusual
nowadays that research teams have access to computation clusters capable of
performing several years of CPU-time in a few weeks. Reviewers may not have
access to such resources, neither the time or the budget required to reproduce
all experiments.

\paragraph{Verification of artifacts}
Although we believe that a cursory peer-review of artifacts before publication
would have a positive impact on reproducibility  in EC, in an ideal world 
one would like to ensure the correctness of the artifacts. However the effort to do so manually 
is tremendous, and can only be reduced somewhat by implementation-agnostic validation testsuites and detailed source documentation. This is also one of the reasons why replicability studies using independent implementations solely based on the description of the algorithm in the paper are very valuable, as it is unlikely that different teams would make exactly the same implementation errors.

\subsection{Obstacles specific to generalisation}

A particular challenge in empirical EC research is that experiments are necessarily limited to specific problem instances, computational budget and parameter settings. Nevertheless, the insights are usually expected to generalise to other settings. For example, if a paper finds one TSP crossover operator superior over another on some TSP instances and for a certain computational budget,
the expectation is (and the claim in the paper usually implies) that similar results also hold for other TSP instances and a larger or smaller computational budget. Studies on generalisability as defined in Table~\ref{tab:terms} are thus very important to understand how robust and generalisable the paper's conclusions are.
Furthermore, in EC, parameter settings can have a huge impact on performance. \citet{SmiEib2010cec} have demonstrated that automatic optimisation of the parameters can substantially improve even the performance of the algorithm winning the CEC 2005 competition. They speculate in their conclusion that different algorithms might benefit differently from tuning, and that tuning all algorithms may change the ranking observed in the competition. This is exactly what \citet{MelDyeBlu2017neural} have investigated for natural language processing. They re-evaluated several popular architectures and regularisation methods by automatically tuning their parameters and arrive at the conclusion that standard architectures, when properly tuned, outperform more recent models. Further evidence is provided by various propositional satisfiability (SAT) competitions~\citep{HutLinBal2017aij} where the rankings of solvers change substantially before and after automatic parameter tuning.
To make things worse, one can argue that changing the problem instance class or the computational budget available necessitates a change in the algorithm's parameter settings, e.g., see \citet{BezLopStu2017assessment}.
Someone testing the generalisability of a paper's conclusion to a different class of problem instances thus faces the additional challenge of choosing appropriate parameter settings.
So unless parameters have been set in a systematic way, one may question whether the observation that one algorithm is better than another is really due to the algorithmic differences, or just a consequence of insufficient or inappropriate tuning.

\section{Guidelines and tools}\label{sec:guidelines}

In this section, we discuss a few general guidelines and present pointers to the literature that  
aim at improving, assessing and encouraging reproducibility of research published in the EC field.
Some of these guidelines are inspired by the ACM guidelines for artifact review badging~\citep{ACM2020badging_v1_1}, the guidelines for AI research endorsed by AAAI\footnote{Latest version can be found at \url{https://folk.idi.ntnu.no/odderik/reproducibility_guidelines.pdf} (Last accessed, version 1.3, June 25, 2020)}~\citep{GunGilAha2018repro}, the Replicated Computational Results
  Initiative of \emph{ACM Transactions on Mathematical Software}~\citep{Her2015toms} and other sources~\citep{StoMcNutBai2016enhancing}.

\subsection{Ensuring reproducibility}
\label{sec:guide-ensuring}

\paragraph{Publish permanently accessible, complete and useful artifacts}
When sharing artifacts, the rule-of-thumb heuristic should be that a person
who only has access to the published paper and the artifacts provided should
be able to reproduce the results shown in the paper without having to contact
the original authors. This implies that the shared artifacts should
not change after publication, because the changes may prevent reproducing the
paper as published. Hence, a development repository, e.g. in GitHub, is not a
valid repository for artifacts unless the precise versions used in a paper are
clearly tagged. Preferably, artifact repositories will have a digital object
identifier (DOI), such as those generated by Zenodo (e.g.
doi:\,\href{http://doi.org/10.5281/zenodo.3749288}{10.5281/zenodo.3749288}). If
revisions to the published artifacts are necessary, it should be easy to
identify each previous version. The repository should provide a plan for
long-term, ideally permanent, accessibility. Authors' personal webpages or
development repositories do not typically satisfy this requirement.  ACM uses
the badge \emph{artifact available} for papers that match the
requirements above~\citep{ACM2020badging_v1_1}.

Artifacts should contain, at a minimum, all the source code and the input data
required to reproduce the results reported in the paper, together with clear
metadata and sufficient documentation on how to reproduce the results. We
suggest, however, to provide a detailed step-by-step documentation, flexible
reproduction scripts and, as much as possible, raw intermediate (generated)
data that allow reviewers and other researchers to selectively repeat parts of
the experiment. Such extensive artifacts enable a better evaluation and
comprehension, hopefully simplifying reproduction efforts and avoiding
mistakes. In summary, with regards to source code, we suggest splitting the
code into:
\begin{itemize}
	\item \emph{Pre-processing code}, e.g. code that generates instance data and 
		scripts that set up the experimental conditions.
	\item \emph{Algorithm code}, the implementation of the algorithm(s) to be
		tested.  
	\item \emph{Analysis code}, scripts that post-process
		the data produced
		by the algorithm and perform statistical analysis.
	\item \emph{Presentation code}, e.g. scripts that generate tables and 
		figures reported in the article.
\end{itemize}

As for the generated data provided, although a paper may report only summary
statistics, the artifacts should ideally contain the raw data generated, thus not only
enabling the reproduction of the analysis, but also further analysis by others.
Furthermore, in an optimisation context, we support the recommendation that the raw data should
contain not only objective function values but also the actual solutions~\citep{GenGraMac1997hownotto,KenBaiBla2016good}, thus
making it possible to verify and compare results. Even for simple problems such as
the TSP, the correct computation of the objective function may depend on
technical details, e.g. preprocessing of distance data.\footnote{E.g.
  \url{http://comopt.ifi.uni-heidelberg.de/software/TSPLIB95/tsp95.pdf}} Subtle
implementation errors cannot be detected unless the actual solutions are
available. Verification may be facilitated by publicly available solution checkers or the
authors themselves may provide such a checker as an additional artifact that
other researchers may use to verify their obtained solutions. Solution
checkers should be as simple as possible so that the implementation can be
trusted. One further step would be to automatically run the solution checker
during post-processing. \citet{GenGraMac1997hownotto} suggest checking every solution evaluated. This proposal will bring us closer to ``certifying
artifacts''~\citep{McConMehNah2011certifying}.

Finally, we argue that artifacts should be made available as source code and
open-data formats under conditions no more restrictive than those required to
read the paper itself. License information should be included with the artifacts, preferably an open-source license allowing reading and distributing the code, running it and,
ideally, reusing it~\citep{StoMcNutBai2016enhancing}. 
 
In the case of papers
using sensitive artifacts that cannot be made available in this manner, we
suggest to perform the experiments supporting the main conclusion using
artifacts that are free from such concerns, either by generating synthetic
data, removing from the source code any sensitive parts or using a less
realistic version of the artifacts, possibly at a different level of
abstraction, as we will discuss later in this section. Results using the
sensitive artifacts may still be reported to highlight qualitative differences,
but they will not constitute the main scientific evidence.

\paragraph{Facilitate access to computational resources.}
Current practice for journals  checking artifacts is that it is up to reviewers to get access to the required resources and bear the cost for reproducing experiments. If special hardware is required, e.g, graphical processing units (GPUs),  authors could consider providing  reviewers with
access to the required resources for the purpose of reproducibility checks. Although
this case may seem similar to the availability of sensitive artifacts discussed
above, where we argued against making sensitive artifacts only
  available to reviewers, there is a fundamental difference: Artifacts that
are only disclosed to reviewers will never become available to other
researchers, which hinders reproducibility, whereas specialist hardware such as
GPUs is publicly available for purchase by interested researchers but reviewers
should not bear the cost. A similar
distinction may be made between undisclosed data, which is not suitable for
reproducibility, and data that is simply too large to host or copy for review
purposes~\citep{FonTag2020repro}. 
Journals might consider making resources available to their reviewers and bear some of the cost. For really expensive resources, however, the only realistic solution might be that research councils specifically fund reproducibility studies (see also Section~\ref{sec:encouraging}).

\paragraph{Report detailed experimental conditions}

Any details required to reproduce the experiment but not included as part of
the artifacts should be thoroughly reported in the documentation included with
the artifacts. These details include the precise versions of any additional
software, packages, libraries, simulators, compilers, interpreters, and
operating systems (possibly including installation steps unless trivial); as
well as the \emph{relevant details} of the hardware platform. For example,
experiments requiring significant amounts of memory should report the memory
available, whereas results sensitive to small changes in computation time
should report full CPU details including cache sizes.

Literate programming,
dynamic documentation and reproducible notebooks\footnote{E.g. Rmarkdown, Jupyter
  notebooks, Knitr} integrate code, documentation and analysis, which makes it much easier to understand and interact with code, and reproduce or observe results by automatically re-creating analysis, tables and figures.

Nowadays, several technical solutions exist that can ensure the portability of 
programs to different software environments such as virtual machines, containers, and platforms, e.g. Open Science Foundation\footnote{\url{https://osf.io/}}, Code Ocean\footnote{\url{https://codeocean.com/}}
and Docker\footnote{\url{https://www.docker.com/}}. 
A container includes everything that is needed to run an application, such as code, system tools, system libraries and settings, independent of the underlying operating system. 
ACM Transactions on Evolutionary Learning and Optimization (TELO) explicitly supports the use of Code Ocean and can integrate it directly into its Digital Library platform.

In the case of algorithms, all (hyper-)parameters should be clearly described
in the documentation, including their domain. For the purposes of generalisability, the process used for setting
parameter values should be reproducible as well, ideally by means of design of
experiments~\citep{Montgomery2012,PaqStuLop07metaheuristics} or automatic
algorithm configuration tools~\citep{Birattari09tuning}, such as
SMAC~\citep{HutHooLey2011lion} or irace~\citep{LopDubPerStuBir2016irace}, with
a clear explanation of the values explored.

If an experiment relies on a random number generator,
one should document or provide as artifacts the precise random seeds that produce the results reported, for
the purpose of allowing the exact repetition of the experiment~\citep{Johnson2002,GenGraMac1997hownotto,KenBaiBla2016good}.\footnote{Of course, the conclusions should not depend on the precise seeds and a reproducibility study should vary the seeds.} Various conferences~\citep{AAAI2021checklist,IJCAI2021checklist} already encourage  the specification of random seeds in their submissions guidelines.
This recommendation  also applies to
randomly generated data and problem instances, although in this case, one may also
provide the data generated for completeness.

\paragraph{Measure and report with reproducibility in mind.}
There are two main concerns that should guide which measurements are performed
and how they are reported: (1) the level of algorithmic abstraction being
considered in the experiment and (2) what measurements can actually be
reproduced given the artifacts provided.

\citet{McG2012guide} provides detailed guidelines for measuring and reporting solution quality and computational effort at various abstraction levels that are directly applicable to EC. At the highest level, we find algorithm paradigms such as metaheuristics, which
    are generic algorithmic templates that can be applied
    to different problem domains; whereas at the lowest level we find executable implementations of specific algorithms running on a particular machine.
    It makes sense to consider machine-independent measures to 
compare algorithm paradigms as well as very precise machine counts to 
compare different implementations, but not the other way around.

When measuring and reporting
computation time, authors should report not only hardware configuration, but
also include calibration codes and their running times among the artifacts provided, e.g. a publicly available deterministic algorithm for the particular problem domain, run on a few small standard benchmark problem instances. Such information may be used to normalise  machine speeds~\citep{Johnson2002} by comparing the speed of these standard benchmarks on different computers, and scaling speeds accordingly.

With respect to ensuring reproducibility in the narrow sense, random experimental
factors may lead to differences in results reported no matter how detailed the
artifacts provided are. Therefore, results should not be reported with a confidence
or precision larger than what can actually be reproduced, since it provides a
false certainty about the values reported. Nevertheless, the more details are
included in the artifacts (e.g. random seeds, precise versions of required
software or even fully-fledged software containers and virtual machines), the
less random variation we need to account for in a reproducibility study.

\paragraph{Report statistical inference to make your claims more robust}
Due to the stochastic nature of EC algorithms,
it is expected that authors report not only means and variances, but also confidence intervals, $p$-values and/or size effects estimates. The usage of confidence intervals, rather than $p$-values, is usually recommended in recent literature~\cite{Cumming2012}.
The former gives useful information about uncertainty and it is easier to interpret. Moreover, a $p$-value can always be derived from a confidence interval, but not the 
other way around.
Effect size, such as Cohen's $d$ and Pearson's $r$ estimate the effect of a treatment, 
such as the effect of a new operator on the overall performance of an algorithm. 
Confidence intervals for the effect size estimates are also available. For an appropriate treatment of inferential procedures in the context of computer science experiments, we refer to~\citet{Cohen1995ai}, \citet{Lilja2000measuring},  \citet{BarChiPaqPre2010emaoa} and \citet{McG2012guide}.

\paragraph{Be precise about the claims made}
Most empirical results in EC are obtained with specific algorithmic parameter
settings, on a small set of problem instances and under specific experimental
conditions (e.g. number of function evaluations allowed) and random
seeds. However, it is usually expected that the conclusions generalise beyond
the precise experiment reported by the paper. Certainly, in most cases, a
conclusion that depends on the specific random seeds would not have much value,
even if it is fully repeatable. On the other hand, the
conclusions in many papers are much broader, e.g. crossover operator A is
better than crossover operator B for continuous optimisation. If a subsequent
study finds that crossover operator B is better than crossover operator A on
continuous problems different than the ones used in the original paper then,
strictly speaking, the original claim is falsified, even though the experiment may still be replicable and we can only  say that the results do not generalise (Table~\ref{tab:terms}).
Authors should therefore be as precise as possible about the claims they make,
such as the experimental conditions and problem classes for which they believe
their conclusions to hold. The experimental design should reflect as well the
scope of the claims, e.g. by using a problem instance generator whenever
possible to clearly define the relevant class of problems rather than testing
on a few arbitrarily selected problem instances. 
In absence of (or in addition
to) such generator, e.g. for complex real-world problems,
defining and measuring problem features would characterise
the scope of the claims~\citep{MunSmi2020ec} and provide evidence that the
conclusions hold within this scope. Another well-known issue is that
specialising algorithm designs and parameter settings to particular problem
instances, i.e. \emph{overfitting}, typically comes at the cost of worsening
performance in unseen instances, even of the same
problem~\citep{Birattari09tuning}. Thus, several
journals~\citep{JoH2015:policy,Dor2016sipolicy} have adopted policies that
require a clear separation between the problem instances used for algorithmic
development and parameter tuning, and problem instances used for hypothesis
testing and benchmarking~\citep{EibJel2002critical}. Such separation provides evidence that the claims of
the paper apply to a broader scope than the particular instances evaluated. The procedure for defining the two sets of instances should be clearly described and reproducible.
Finally, a sensitivity analysis of parameter settings and experimental
conditions would also provide evidence that the main conclusions hold when
those conditions vary.

\subsection{Assessing Reproducibility}

Procedures to assess reproducibility should be tied to author's claims. Unfortunately, there is no standard to evaluate reproducibility of 
computational experiments. This is particularly difficult in EC, 
since one has to deal not only with differences on hardware and/or 
software for reproducing experiments, but also with algorithm stochasticity. 
Therefore, we advocate some caution before concluding,
in a clear-cut manner, that a work is not reproducible if results do not match exactly. Instead, we suggest  to investigate further 
reasons for the work not being reproducible, for instance, identify
possible hardware or software differences, such as compiler flags, 
cache level sizes, software libraries, or even sample size.

Even in repeatability studies (see Table~\ref{tab:terms}), we may not always expect an implementation to obtain exactly the same result under the same random seed if run twice, for instance, due to small fluctuations on the running time that defines the termination condition. Inferential procedures could be used to assert whether the  differences between
the original runs and the replicated runs are due to a 
random or a systematic effect. In this case, a matched-pair
inferential procedure would be appropriate in order to take into account  the natural pairing between the original and the replicate run using the same random seed.

A typical scenario in EC is to compare the performance of several algorithms on a
set of benchmark instances. Asserting an author's claim that Algorithm A is significantly better than Algorithm B with respect to solution quality can be performed by testing whether the reproduced results show a significant effect in the same direction, given the same significance level as specified in the original publication. Alternatively, the opposite direction of the claim could be tested, which, if significant, would allow to infer that the work is not reproducible.

In the above scenario, it is also possible to test if the effect 
size is significantly different, even if the direction is the same.
The authors in \citet{OSC2015estimating} suggest to test whether the original 
effect size is within the 95\% confidence interval of the reproduced effect size estimate.
However, some concerns have  been raised about this procedure, 
as the average probability of the first 95\% confidence interval including the next reproduced mean is only approximately 83\%~\cite{Cumming2012}. Reporting confidence intervals of the difference between original and reproduced effect sizes is usually recommended. If 0 is included in this confidence interval, it suggests that the work is reproducible. 

We note that the usual assumptions of parametric inferential 
procedures may be hard to be met for assessing EC algorithms and
non-parametric alternatives may be better suited. However, conducting non-parametric inference procedures based on 
computationally intensive methods, such as bootstrapping and randomisation tests, for assessing reproducibility may require access to all data collected by the original study in addition to the aggregated statistical measures usually reported.

\emph{Meta-analysis} is an interesting complementary analysis extensively used in other fields to aggregate results from different studies to derive general conclusions
\cite{BorHedHigRot2009metanalysis}. A recent example is the meta-analysis of the effect of adaptiveness in adaptive large neighborhood search~\cite{TurSorHva2021meta}.
In the context of reproducibility, meta-analysis would allow to
understand how much the effect size varies in the 
original and the reproduced studies by combining the results from 
both. This new estimate takes usually the form of a weighted average
of individual estimates, where the weights are inversely proportional 
to the sampled variance, and from which inferential 
procedures for testing heterogeneity are constructed.
We refer to \citet{Ehm2016} for the application of inferential 
methods in meta-analysis for reproducibility. %

Most research in EC is trying to derive general insights such as ``algorithm A is better than algorithm B for the class of problems with feature C'' from limited experiments on a set of problem instances.
Reproducibility studies should thus not only focus on exactly reproducing results, but also expand experimentation to assess generalisability, by changing the value of fixed factors such as parameter settings and  problem instances. Statistical methods exist to assess the generality of conclusions even from a limited number of real-world problem instances~\citep{Bar2015genera}.
Such experiments will  confirm, over time, that the  conclusions are not only valid for the fixed values  examined in the original paper, but have a broader validity.

\subsection{Encouraging reproducibility efforts\label{sec:encouraging}}

Ideally, rigorous journals should adopt the Transparency and Openness Promotion
(TOP) guidelines, which require reproducibility checks and even independent
replication before
publication~\citep{StoMcNutBai2016enhancing,NosAltBank2015promoting}.
Some journals, such as \emph{Mathematical Programming Computation}\footnote{\url{https://www.springer.com/mathematics/journal/12532} (Last accessed: January 22, 2021).},  already require that source code is provided to reviewers and we concur with authors who argue that  this requirement should become the norm~\citep{SorArnPal2017}.
Top conferences in Artificial Intelligence have recently adopted  reproducibility checklists as part of their submission process, e.g. NeurIPS~\citep{PinVinSin2020improving,PinSin2020checklist}, AAAI Conference on Artificial Intelligence~\citep{AAAI2021checklist} and  International Joint Conference on Artificial Intelligence~\citep{IJCAI2021checklist}.
An intermediate, less onerous step is to award recognition to papers that
achieve certain levels of reproducibility. ACM badges
\citep{ACM2020badging_v1_1} already provide a way to recognise different
degrees of reproducibility that journals could adopt.

ACM Transactions on Evolutionary Learning and Optimization (ACM TELO) follows the ACM guidelines for
reproducibility~\citep{ACM2020badging_v1_1}. When submitting the manuscript, the author can 
apply for an ACM reproducibility badge. Once the paper passes the first stage of review and is accepted or returned to the author for revision, the artifacts are reviewed by a member of the journal's reproducibility board, who can recommend that a badge be awarded, or request further revisions of the artifact before a badge can be awarded. Three badges can be requested: \emph{Artifacts Available}, \emph{Artifacts Evaluated} and
\emph{Results Reproduced}. The badges are independent, that is, any combination
of badges can be requested. The badge \emph{Artifacts Available} is provided if
the artifact is publicly available in a permanent repository.  The badge
\emph{Artifacts Evaluated} is concerned with ensuring that the artifact
fulfills the requirements to be reproduced by others and it has two levels,
\emph{functional} and \emph{reusable}, the latter requiring that the artifact
can be re-used and re-purposed.  The badge \emph{Results Reproduced}
corresponds to the notion of reproducibility presented in Section~\ref{sec:terminology}, that is,
the experimental results are validated using the artifact provided by the
author.  In order to receive this badge, it is required that the results
obtained by the reviewer are in agreement with those in the article  within a
tolerance deemed acceptable. For this reason, it is required that precise
estimates of performance are reported by the authors.
Note that ACM also has a badge for \emph{Results Replicated}, which requires that the results are replicated without using the author's artifacts. However, this is not offered by ACM TELO at the moment, as the effort to replicate the code has been deemed excessive.

Another aspect that may be of interest to the EC community and that may help to
prevent publication bias, is to allow authors to
\emph{pre-register}~\citep{NosEbeHav2018preregistration}
their scientific
studies and hypotheses with a journal, or a publicly available website, before
conducting the experiments. Pre-registration would allow reviewers to verify
whether the initial authors' plan and the published results match or not.  A
more ambitious goal would be to allow the pre-registration document to be peer
reviewed in order to identify issues with the experimental setup and its
suitability for validating the authors' hypotheses before the experiments are
conducted.  A certain publication guarantee could be provided depending of the
reviewers' confidence.  This is particularly relevant for experimental studies
that take very long time or require huge amounts of computational resources.

Funding agencies may also encourage reproducibility in various ways, as
suggested by \citet{StoMcNutBai2016enhancing}. In particular, funding agencies
may require that the resulting research is reproducible according to specific
and verifiable criteria, in a similar manner that some funding agencies already
require and/or provide funding for open-access publications and data management plans.
Funding agencies could also encourage and support reproducibility efforts by funding reproducibility studies as well as research that analyses or alleviates
reproducibility obstacles.  We want to highlight the incongruity of funding
non-reproducible research with public money.  %

\section{Discussion and conclusions}\label{sec:conclusions}

Reproducibility is a cornerstone of science. Without reproducibility, scientific progress is impossible. Yet, many scientific works are not reproducible.
EC is particularly vulnerable because of its reliance on experimental results and the stochastic nature of its algorithms. In this paper, we have discussed reproducibility in the context of EC, and proposed a new classification of reproducibility studies, distinguishing four different types, namely, \emph{Repeatability}, \emph{Reproducibility}, \emph{Replicability} and \emph{Generalisability}, with different purposes and study designs. Our proposed classification could be applied to other fields in Computer Science. We have then analysed the reasons for the reproducibility crisis and identified various cultural and technical obstacles.

Despite these obstacles, there are positive developments that point to a shift of
culture. First, concern is growing in the EC community about questionable
benchmarking~\citep{EibJel2002critical,BarDoeBer2020benchmarking}, insufficient statistical assessment~\citep{GarMolLoz2009joh,ShiMarDud2008stat,Buz2019signif}, unfair
parameter tuning~\citep{Birattari09tuning,BarPre2014experimental}, and, more recently, reproducibility
and replicability issues~\citep{KenBaiBla2016good,SorArnPal2017}.
Several journals have adopted explicit policies that
encourage reproducibility\hspace{0pt}---albeit do not require it---and improve
replicability. Some ACM journals, with TELO being a prominent example, have
established reproducibility boards that award badges recognising the effort in
making research reproducible. Finally, due to this
shift in culture, solutions to technical obstacles are becoming more widely
available and adopted, thus lowering the effort to improving the
reproducibility  of EC research.

We suggest that reproducibility (in the narrow sense) is a short-term goal that
ideally should be checked during the review process. In EC, in particular,
there are no actual technical obstacles to make code and data available, thus
making results reproducible should be the norm. Platforms such as CodeOcean and
OSF exist that provide nearly identical experimental setup to ensure that
published results may be reproduced by reviewers and other
researchers. Nevertheless, once this validation step is done, we believe that
the preservation of code and data is more useful in the long-term than the
long-term availability of a reproducible experimental environment, given the
rapid obsolesce of software and hardware. Even if the original study becomes
non-reproducible due to the obsolescence of its original artifacts, studying
their code and data could help future replication and generalisation efforts.

The next step should be empirical and statistical replicability, and published
research should enable it. In other words, published research should contain
the information required to independently replicate the experiment without
using the original artifacts, and reach the same conclusion given the
statistical confidence claimed by the original experiment. This information
would include all relevant details about the algorithm, problem, measurements
and experimental environment at the \emph{right} level of abstraction. It would
also include all statistical details that would allow the authors of a
replication study to assess whether their new results, which are expected to be
numerically different from the original ones due to varying the random factors
of the experiments, reject or not the original hypothesis.  The final step that
will actually push the boundary is to examine the generalisability of the
claims made in scientific papers by testing whether the main conclusions still
hold in somewhat different experimental setups and for different problem
classes.

To overcome the reproducibility crisis we need a culture shift towards reproducibility in EC, with reproducibility playing a bigger role
in education, funding decisions, recruitment and reputation. While this requires some extra effort, especially early on, the
reward will be faster scientific progress, less frustration trying to build on other's work, and a higher reputation for the field as a whole. The journey has already begun.

\begin{acks}
We would like to thank Carola Doerr (Sorbonne University, France) and Mike Preuss (Leiden
University) for pointing
out guidelines for reproducibility in other fields.
M.\@ L\'opez-Ib\'a\~nez is a ``\emph{Beatriz Galindo}'' Senior Distinguished Researcher (BEAGAL 18/00053) funded by the Spanish Ministry of Science and Innovation (MICINN).
  This work was partially funded by national funds through the FCT - Foundation for Science and Technology, I.P. within the scope of the project CISUC -- UID/CEC/00326/2020.

\end{acks}

\providecommand{\MaxMinAntSystem}{{$\cal MAX$--$\cal MIN$} {Ant} {System}}
  \providecommand{\rpackage}[1]{{#1}}
  \providecommand{\softwarepackage}[1]{{#1}}
  \providecommand{\proglang}[1]{{#1}}

\end{document}